\documentclass[10pt,twocolumn,letterpaper]{article}

\usepackage{etoolbox}
\newtoggle{arxiv}
\toggletrue{arxiv}
\iftoggle{arxiv}{
    \usepackage[pagenumbers]{wacv}
}{
    \usepackage{wacv}
}
\usepackage{float}
\usepackage{multirow}
\usepackage{listings}
\usepackage{colortbl}
\usepackage{lipsum}
\usepackage{cuted}
\usepackage{pifont}
\usepackage{subcaption}
\usepackage{xcolor}

\def\OURS{CVP\xspace}
\def\grid{allocentric}
\def\token{target-affinity}
\def\Grid{Allocentric}
\def\Token{Target-affinity}

\title{\OURS: Central-Peripheral Vision-Inspired Multimodal Model for Spatial Reasoning}

\definecolor{wacvblue}{rgb}{0.21,0.49,0.74}
\usepackage[pagebackref,breaklinks,colorlinks,allcolors=wacvblue]{hyperref}

\author{
Zeyuan Chen$^1$ \quad
Xiang Zhang$^1$ \quad
Haiyang Xu$^1$ \quad
Jianwen Xie$^2$ \quad
Zhuowen Tu$^1$ \\
$^1$UC San Diego \quad $^2$Lambda, Inc.
}

\begin{document}

\twocolumn[{%
    \renewcommand\twocolumn[1][]{#1}%
    \maketitle
    \begin{center}
    \includegraphics[width=\linewidth]{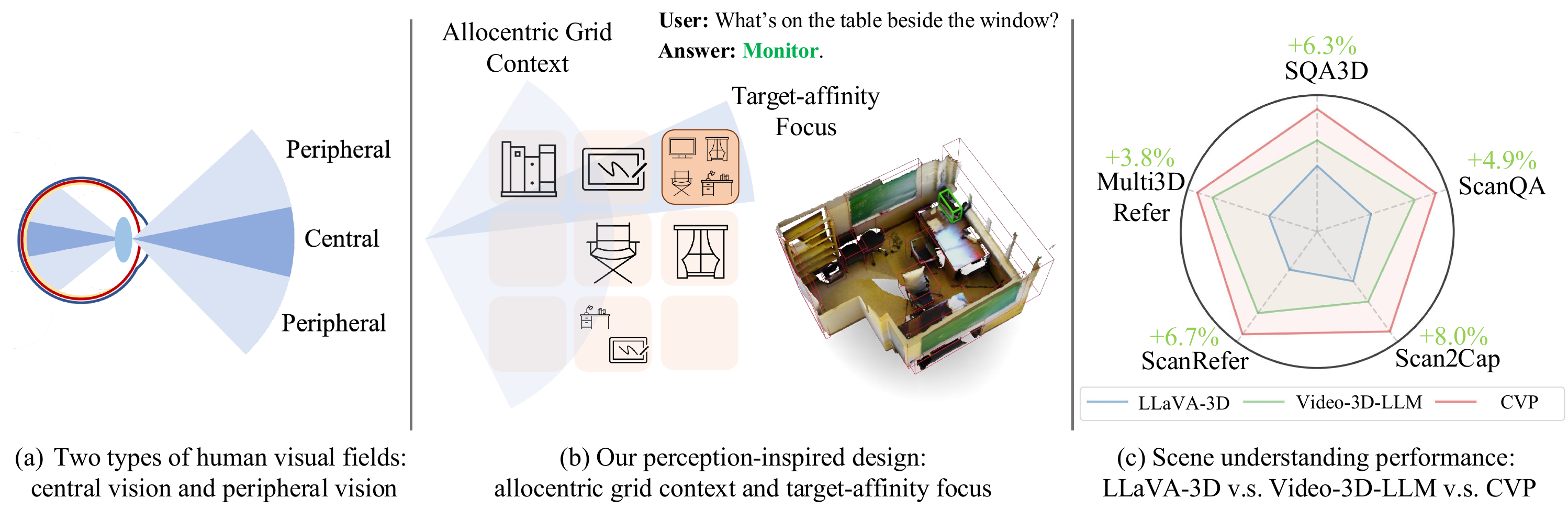}
    \captionof{figure}{Overview of \OURS{}, inspired by human visual cognition. (a) Human vision combines central vision (dark blue) for focused, high-acuity perception and peripheral vision (light blue) for broader contextual awareness. (b) Our model mimics this dual process with \textbf{\token{} token}, guiding attention to target-relevant objects/regions, and \textbf{\grid{} grid}, capturing allocentric spatial context. (c) Quantitative results across multiple 3D scene understanding benchmarks. We report EM on SQA3D~\cite{sqa3d}, CIDEr on ScanQA~\cite{scanqa} and Scan2Cap~\cite{scan2cap}, Acc@0.25 on ScanRefer~\cite{scanrefer}, and F1@0.25 on Multi3DRefer~\cite{multi3drefer}. With \token{} token and \grid{} allocentric grid, \OURS{} consistently outperforms state-of-the-art models such as LLaVA-3D~\cite{llava3d} and Video-3D-LLM~\cite{video3dllm}. }
    \label{fig:teaser}
    \end{center}
    \vspace{20pt}
}]

\begin{abstract}
We present a central-peripheral vision-inspired framework (\OURS{}), a simple yet effective multimodal model for spatial reasoning that draws inspiration from the two types of human visual fields -- central vision and peripheral vision. Existing approaches primarily rely on unstructured representations, such as point clouds, voxels, or patch features, and inject scene context implicitly via coordinate embeddings. However, this often results in limited spatial reasoning capabilities due to the lack of explicit, high-level structural understanding. To address this limitation, we introduce two complementary components into a Large Multimodal Model-based architecture: \textbf{\token{} token}, analogous to central vision, that guides the model's attention toward query-relevant objects; and \textbf{\grid{} grid}, akin to peripheral vision, that captures global scene context and spatial arrangements. These components work in tandem to enable structured, context-aware understanding of complex 3D environments. Experiments show that \OURS{} achieves state-of-the-art performance across a range of 3D scene understanding benchmarks.
\end{abstract}    
\section{Introduction}

Understanding 3D scenes is a fundamental ability that agents need to have to interact with the physical world, enabling applications in robotics~\cite{thrun2002probabilistic, newcombe2011kinectfusion, salas2013slam++, he2025asap}, autonomous navigation~\cite{voxelnet, pointpillars, frustumpointnet, min2024driveworld, muhammad2022vision}, and embodied AI~\cite{embodied1, embodied2, embodied3, embodied4}. Common tasks for 3D spatial understanding, such as 3D question answering (\eg ScanQA~\cite{scanqa}, SQA3D~\cite{sqa3d}), 3D object localization (\eg ScanRefer~\cite{scanrefer}, Multi3DRefer~\cite{multi3drefer}) and dense captioning (\eg Scan2Cap~\cite{scan2cap}) all require a model to jointly reason about geometry, semantics, and spatial relationships.

The progress of Large Language Models (LLMs)~\cite{gpt3,gemini,llama,llama2,deepseek,achiam2023gpt} and Large Multimodal Models (LMMs) for images~\cite{llava, gemini,achiam2023gpt,gemini15} and videos~\cite{llavavideo,llavaonevision,qwen25, achiam2023gpt} has catalyzed advances in spatial intelligence. To extend these models to 3D scene understanding, one line of work~\cite{3dllm,ll3da} integrates scene-level geometric representations, such as point clouds, voxels, or patch features, into LLMs via lightweight adapters, including LLaVA~\cite{llava}-style projection layers or cross-attention-based modules like Q-Former~\cite{qformer}. These methods retain dense geometric detail and broad spatial coverage. However, mapping unordered 3D features into the token space of language models presents challenges: correspondences are often noisy, positional cues are injected implicitly through coordinates, and training can produce weak vision–language alignment due to the limited availability of labeled 3D data. Another direction~\cite{chat3dv2,leo,chatscene} follows the egocentric perception of embodied agents, leveraging object-centric representations wherein object proposals are enriched with associated 2D and 3D features. This structure improves semantic grounding and reduces token load, but often fragments global context and under-represents inter-object spatial relationships. Most recently, researchers~\cite{llava3d,video3dllm} have explored using pre-trained LMMs by sampling videos present in 3D datasets and feeding the resulting frame sequences to the model. While these models achieve promising performance on various benchmarks, 3D information is typically injected only implicitly, \eg via coordinate embeddings, leaving the model without an explicit, high-level allocentric representation of scene structure.

To address these limitations, we draw inspiration from the dual nature of the human visual field. Human vision consists of a high-acuity center region (fovea) and a low-acuity peripheral region. This foveal-peripheral structure allows humans to focus sharply on regions of while maintaining a broad awareness of the surrounding environment. Inspired by this efficient perceptual mechanism, several computer vision models have explored similar designs for tasks such as scene parsing~\cite{li2017foveanet}, navigation~\cite{thavamani2021fovea}, and saliency  detection~\cite{recasens2018learning}. Mimicking this mechanism can help models allocate resources more effectively, prioritizing relevant details without losing sight of the global scene.

Motivated by this insight, we propose \OURS{}, a framework that incorporates the following two key components designed to enhance the 3D spatial understanding ability of the base LMM model:

(1) \textbf{\Token{} token}, which directs the model’s attention toward query-relevant objects, serving a role analogous to human central vision. This mechanism is trained with a contrastive loss, where positive samples correspond to target-relevant objects.

(2) \textbf{\Grid{} grid}, which captures the global scene layout from a world-centric (allocentric) perspective, akin to peripheral vision. Object locations are discretized into a bird’s-eye-view grid, enabling structured reasoning over spatial relationships.

These two components enable \OURS{} to reason effectively over both fine-grained object details and global spatial context. Combined with the strong pretraining of the LMM, \OURS{} achieves state-of-the-art performance across five representative 3D scene understanding benchmarks, as illustrated in \cref{fig:teaser} (c). For example, it achieves 62.3 EM on SQA3D~\cite{sqa3d}, 107.1 CIDEr on ScanQA~\cite{scanqa}, 62.0 Acc@0.25 on ScanRefer~\cite{scanrefer}, and 60.2 F1@0.25 on Multi3DRefer~\cite{multi3drefer}, surpassing the previous state-of-the-art Video-3D-LLM by $+$3.7 EM, $+$5.0 CIDEr, $+$3.9 Acc@0.25, and $+$2.0 F1@0.25, respectively.

Our contributions are summarized as follows:

\begin{itemize}

\item We propose \OURS{}, a novel framework for 3D spatial understanding, built upon the LMM model and inspired by principles of human visual perception.

\item We design two complementary components to enhance spatial reasoning: (1) \textbf{\token{} token}, which directs the model’s attention to query-relevant objects, akin to central vision; and (2) \textbf{\grid{} grid}, which captures the global scene layout and spatial relationships from a world-centric perspective, resembling peripheral vision.

\item Extensive experiments demonstrate that \OURS{} achieves state-of-the-art performance on five representative 3D vision-and-language benchmarks: ScanQA, Scan2Cap, SQA3D, ScanRefer, and Multi3DRefer.

\end{itemize}

\section{Related Work}
\subsection{3D Scene Understanding}
3D scene understanding is a fundamental problem in computer vision, with critical applications in robotics~\cite{thrun2002probabilistic, newcombe2011kinectfusion, salas2013slam++} and autonomous driving~\cite{voxelnet, pointpillars, frustumpointnet}. With the rise of large language models and large multimodal models, there has been growing interest in integrating 3D perception with natural language understanding~\cite{pointclip, pointbert}. Several vision-language tasks have emerged to benchmark 3D scene understanding. In particular, 3D dense captioning~\cite{scan2cap, more, xtrans2cap} requires models to accurately localize objects and produce descriptions with accurate geometry and semantic information. 3D question answering~\cite{scanqa, sqa3d, parelli2023clip} challenges models to interpret spatial and relational cues in the scene to answer complex queries. 3D visual grounding~\cite{scanrefer,multi3drefer,mvt,multiviewground,3drpnet} requires models to localize objects based on free-form referring expressions. These tasks demand the joint reasoning of semantics, geometry, and spatial relationships. Some approaches utilize Bird’s Eye View (BEV) features for 3D grounding and VQA tasks~\cite{fan2024navigation,wang2025g3d}. \OURS instead adopts an \grid{} grid, which is an abstract textual representation of the 3D scene that allows the language model to more effectively capture the global scene context without introducing modality misalignment.
While expert models have been developed to address each of these tasks individually~\cite{xtrans2cap,butddetr,scanqa,man2024situational,achlioptas2020referit3d,vil3drel,multiviewground,eda}, unified frameworks that can handle all tasks with a single model have long been sought after. The emergence of LMMs provides a promising foundation for such generalist models, offering strong language reasoning capabilities that can be coupled with 3D scene features. Although specialized methods excel in isolated settings, there remains a clear need for a unified model capable of general-purpose 3D vision-language understanding, which is the goal we aim to achieve. 

\begin{figure*}[htbp]
    \centering
    \includegraphics[width=\linewidth]{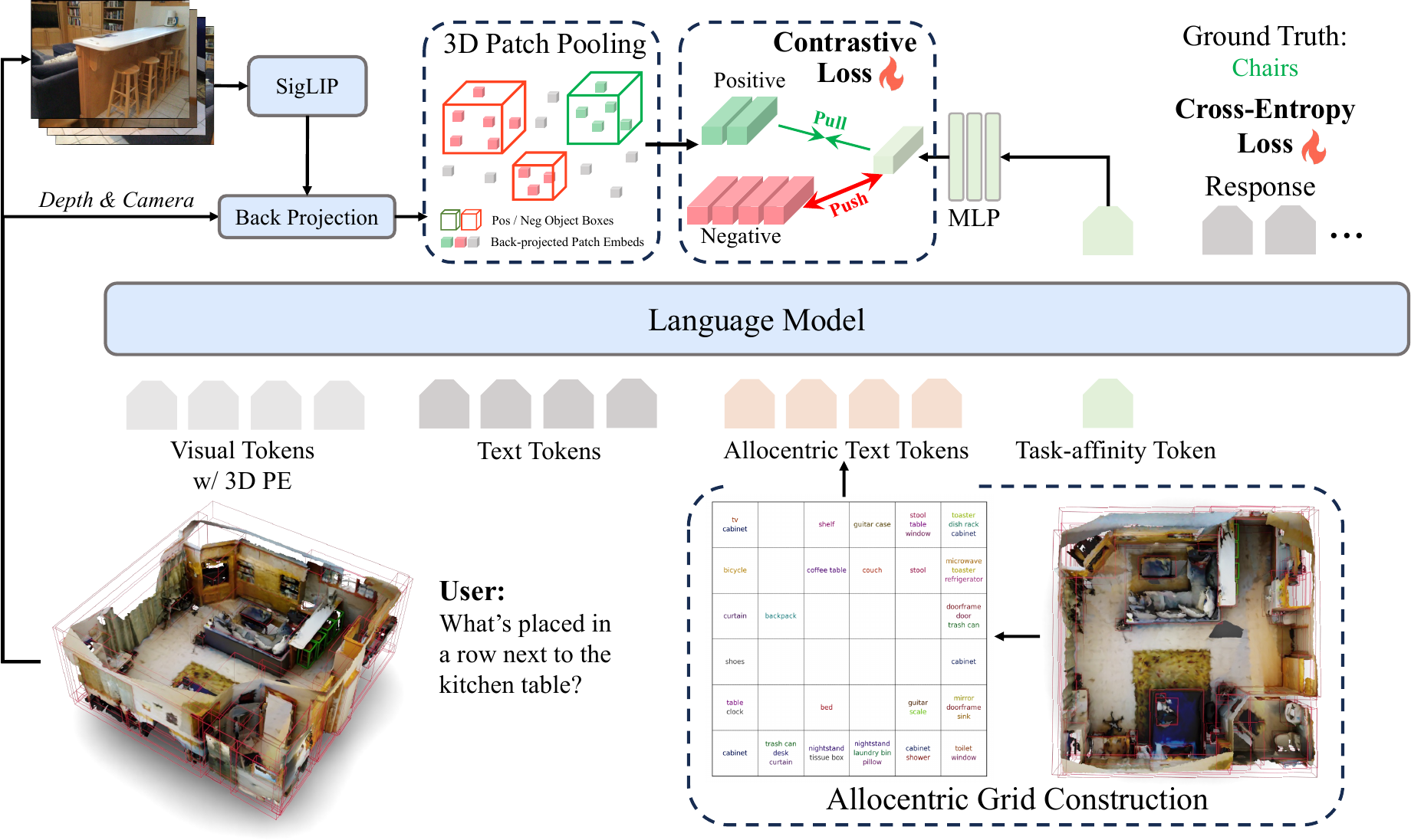}
    \caption{Illustration of \OURS{}. Given visual tokens from multi-view images with 3D positional embeddings and a user question as input, we (1) incorporate a text-based \textcolor[HTML]{D88C70}{\textbf{\grid{} grid}} to provide allocentric global scene context; and (2) introduce a special \textcolor[HTML]{9FCC8C}{\textbf{\token{} token}} that guides the model to focus on target-related objects. During output generation, in addition to producing a language response, the representation of the \token{} token is passed through an MLP and optimized with a contrastive loss against 3D object embeddings back-projected from multi-view 2D features. Positive samples correspond to ground-truth objects relevant to the question, while negatives are irrelevant. This contrastive supervision helps the model attend more effectively to semantically relevant targets.}
    \label{fig:arch}
\end{figure*}

\subsection{Large Models in 3D Scene Understanding}
Recent work has demonstrated that Large Language Models (LLMs) and Large Multimodal Models (LMMs) can be extended to 3D vision–language tasks by aligning 3D representations with textual semantics. PointLLM~\cite{pointllm} re-trains Point-BERT~\cite{pointbert} to extract 3D embeddings and learns a projection from point-cloud features into the language space. LL3DA~\cite{ll3da} adapts the Q-Former architecture~\cite{qformer} to distill point-cloud features into a set of latent tokens, which are then aligned with language via a linear mapping. Object-centric approaches~\cite{grounded3dllm,chat3dv2,leo,chatscene} utilize 3D object bounding boxes to derive instance-level embeddings, which are mapped into the language space using learned projection layers~\cite{grounded3dllm,chat3dv2,leo,chatscene} and contrastive objectives~\cite{grounded3dllm}. While these methods benefit from object-level semantics and category priors, they may struggle to capture holistic, scene-level spatial relationships. Another line of works~\cite{3dllm,llava3d, video3dllm} renders 3D scenes into multi-view 2D images and leverages pre-trained image encoders, such as CLIP~\cite{clip}, to extract multi-view features as a proxy for 3D scene representation. For example, 3D-LLM~\cite{3dllm} aggregates multi-view features into 3D via 3D geometric cues; LLaVA-3D~\cite{llava3d} back-projects image features into 3D space and peforms 3D patch pooling; and Video-3D-LLM~\cite{video3dllm} treats rendered images as video frame sequences and injects 3D spatial information via 3D position encodings. \OURS{} builds upon this multi-view rendering paradigm but introduces two key innovations inspired by human visual perception. At the input stage, it integrates allocentric contextual cues through a symbolic \grid{} grid. At the output stage, it applies contrastive supervision via the \token{} token aligned with target objects. Together, these enhancements promote a more structured and comprehensive understanding of complex 3D environments.

\section{Method}

Human visual perception operates through two complementary components: central vision, which enables high-acuity focus on regions of interest, and peripheral vision, which provides broader contextual awareness across the visual field. Inspired by this dual-process mechanism, as illustrated in \cref{fig:arch}, we propose \OURS{}, a 3D scene understanding framework that incorporates two key modules: (1) \textbf{\token{} token}, which functions analogously to central vision by directing the model’s attention toward question-relevant regions (\cref{sec:token}); and (2) \textbf{\grid{} grid}, which mimics peripheral vision by encoding global scene context and spatial layout from a world-centric perspective (\cref{sec:grid}). We describe the overall training objective used to jointly optimize these components in \cref{sec:t_obj}.

\subsection{\Token{} Token}
\label{sec:token}

In complex 3D scenes, objects are often densely populated. However, for downstream tasks such as 3D question answering or visual grounding, only a small subset of these objects is typically relevant to the query. The presence of irrelevant objects can introduce semantic noise, distract model attention, and degrade model performance. To mitigate this, we introduce a training mechanism that explicitly teaches the base model to learn focusing on target-relevant regions by incorporating a special supervision token, referred to as the \textbf{\token{} token}.

The \token{} token is designed to learn the semantic affinity between the objects relevant to input question and objects in the scene. During training, it is supervised using a contrastive objective that encourages alignment with the embeddings of relevant (positive) objects while pushing away embeddings of irrelevant (negative) ones. This mechanism serves as a lightweight, plug-in module that enhances the model's attention distribution, without requiring additional annotations at inference time.

\subsubsection{3D Object Embedding Extraction}

To enable fine-grained affinity modeling between individual objects, we first extract per-object 3D embeddings. Given a multi-view input tensor with $V$ images $X \in \mathbb{R}^{V \times 3 \times W \times H}$, we adopt SigLIP~\cite{zhai2023sigmoid} as the vision encoder to obtain 2D features maps:
\begin{equation}
    {X}' = \mathrm{SigLIP}({X}), 
    \quad 
    {X}' \in \mathbb{R}^{V \times C \times W' \times H'},
\end{equation}
where $C$, $W'$, and $H'$ denote the feature channel, width, and height, respectively. Each 2D feature map ${X}'_i$ from view $i$ is then back-projected into the shared 3D coordinate space using the known camera intrinsics ${K}_i$ and extrinsics $[{R}_i\mid {t}_i]$. This back-projection operation is consistent with the geometric procedures used in 3D reconstruction systems~\cite{dahnert2021panoptic,zhang2023uni,zhao2025depr}. Specifically, For each pixel location $(u, v)$ with depth $d$, the corresponding 3D point coordinate $\mathbf{p} \in \mathbb{R}^3$ is computed as:
\begin{equation}
    {p} = {R}_i^{-1}\big( d \, {K}_i^{-1} [u, v, 1]^\top - {t}_i \big).
\end{equation}

The feature vector at $(u,v)$ from view $i$ is assigned to its corresponding 3D point $p$. By aggregating projected features from all pixels across all $V$ views, we obtain a 3D point cloud feature set $P$.

For each object $j$ defined by its 3D bounding box, we select the subset of points from $P$ that lie inside the box. The per-object embedding $\mathbf{E}_i$ is then computed via average pooling over the features of these points.

\subsubsection{Contrastive Learning with \Token{} Token}

To train the model to identify target-relevant objects, we introduce a special \token{} token to the input of the LLM. During decoding, the model produces a hidden embedding corresponding to this token, and we denote it as $z_{\texttt{aff}}$. Finally, $z_{\texttt{aff}}$ is passed through a multi-layer perceptron $g(\cdot)$ to produce the query vector $q$:
\begin{equation}
q = g(z_{\texttt{aff}}).
\end{equation}

We then apply the InfoNCE contrastive loss~\cite{oord2018representation}, treating embeddings of target-relevant objects as positives and those of irrelevant ones as negatives. Let $\mathcal{E}_{+}$ denote the set of positive (question-relevant) object embeddings and $\mathcal{E}$ the set of all objects. The contrastive objective is defined as:
\begin{equation}
\mathcal{L}_{\mathrm{InfoNCE}} = - \log \frac{
    \sum_{e_+ \in \mathcal{E}_+} \exp\left( {q^\top e_+} / \tau \right)
}{
    \sum_{e \in \mathcal{E}} \exp\left( q^\top e / \tau \right)
},
\end{equation}

where $\tau$ is a temperature hyperparameter. This objective encourages the query embedding $q$ to be closer (in dot-product similarity) to target-relevant object embeddings $\mathcal{E}_+$, while pushing it away from negative ones $\mathcal{E}_-$. Details of how we obtain the set of question-relevant object $\mathcal{E}_{+}$ are in the supplemental material.

Through this contrastive supervision, the \token{} token learns to capture the semantic alignment between the question and relevant objects. At inference time, the model with the task-affinity token can shift its attention toward question-aligned content without further supervision.

\subsection{\Grid{} Grid}
\label{sec:grid}

Understanding the spatial layout of a 3D scene requires not only recognizing individual objects but also reasoning about their spatial relationships, particularly in the vicinity of query-relevant targets. To support this capability, we introduce the {\grid{} grid}, an allocentric (\ie, world-centered) spatial representation that encodes the local neighborhood structure around each object of interest.

In contrast to low-level dense semantic features that encode appearance, the \grid{} grid emphasizes high-level spatial context, specifically the positions of nearby objects. This representation enables the model to reason about the environment from an allocentric perspective.

We begin by extracting 3D bounding boxes for all objects in the scene and projecting them onto a 2D bird's-eye-view (BEV) plane. To focus on spatial relationships rather than exact object dimensions, we simplify each object to its BEV center point, discarding size and orientation.

The BEV plane is then partitioned into a uniform grid of discrete cells, each corresponding to a fixed spatial region in the 3D scene. For every object, we identify the cell containing its center and record its category name into the set associated with that cell. This yields a sparse, symbolic world-centric grid, where each cell contains a (possibly empty) set of nearby object names.

To inject this structured spatial context into the model, we convert the \grid{} grid into a natural language description. This design ensures full compatibility with the model’s input modality and eliminates the need for additional projection or alignment modules. Specifically, the symbolic contents of the \grid{} grid are serialized into a concise spatial summary that expresses the arrangement of objects across grid cells. Please see the supplemental material for the template prompt and other details.

This lightweight textual encoding enables the model to leverage spatial priors while maintaining simplicity and flexibility. Moreover, it complements other input modalities (e.g., RGB images or depth) by enriching the context with high-level allocentric relationships.

\subsection{Training Objectives}
\label{sec:t_obj}

The overall training objective of \OURS{} consists of two components:

\begin{itemize}
    \item \textbf{Language modeling loss} $\mathcal{L}_{\mathrm{LM}}$, which follows the standard next-token prediction objective for supervising answer generation.
    \item \textbf{Contrastive relevance loss} $\mathcal{L}_{\mathrm{InfoNCE}}$, described in \cref{sec:token}, which encourages the \token{} token to align with embeddings of question-relevant objects while distancing itself from irrelevant ones.
\end{itemize}

The final loss is a sum of the two:
\begin{equation}
\mathcal{L}_{\mathrm{total}} = \mathcal{L}_{\mathrm{LM}} + \mathcal{L}_{\mathrm{InfoNCE}}.
\end{equation}
This objective jointly optimizes both answer generation and object relevance grounding, enabling the model to reason accurately in complex 3D environments.

\section{Experiments}
\subsection{Experimental Setup}
\subsubsection{Datasets}
We evaluate our model on five 3D scene understanding benchmarks. For 3D visual grounding, we use ScanRefer~\cite{scanrefer} and Multi3DRefer~\cite{multi3drefer} to assess the models' object localization capacity given natural language descriptions. For 3D question answering, we incorporate ScanQA~\cite{scanqa} and SQA3D~\cite{sqa3d}, which evaluate spatial and situated reasoning capabilities in 3D scenes, respectively. For 3D dense captioning, we include the Scan2Cap~\cite{scan2cap} benchmark, where the model receives a 3D scene and a target 3D bounding box and is tasked with generating a textual description of the box's contents. In evaluation, we follow the standard protocol adopted in prior works~\cite{grounded3dllm,chat3dv2,llava3d,video3dllm}, using the validation sets for Scan2Cap, ScanQA, ScanRefer, and Multi3DRefer, and the test set for SQA3D.

\subsubsection{Baselines}
To demonstrate the effectiveness of \OURS{}, we compare it with a comprehensive set of baselines across three categories: expert models, 2D large multimodal models (LMMs), and 3D LMMs.

\noindent \textbf{Expert models.} For 3D visual grounding on ScanRefer and Multi3DRefer, we use baselines including ScanRefer~\cite{scanrefer}, MVT~\cite{mvt}, ViL3DRel~\cite{vil3drel}, BUTD-DETR~\cite{butddetr}, 3DVG-Trans, 3DJCG~\cite{3djcg}, and M3DRef-CLIP~\cite{multi3drefer}. For question answering on ScanQA and SQA3D, we compare against Scan2Cap~\cite{scan2cap}, ClipBERT~\cite{clipbert}, MCAN~\cite{mcan}, ScanQA~\cite{scanqa}, and 3D-VisTA~\cite{3dvista}. For dense captioning on Scan2Cap, we include Scan2Cap~\cite{scan2cap}, 3DJCG~\cite{3djcg}, 3D-VLP~\cite{3dvlp}, 3D-VisTA~\cite{3dvista}, and Vote2Cap-DETR~\cite{vote2cap} as baselines.

\noindent \textbf{2D LMMs.} We evaluate against representative large multimodal models including VideoChat2~\cite{videochat2}, Qwen2.5-VL-7B~\cite{qwen25}, and LLaVA-Video~\cite{llavavideo}.

\noindent \textbf{3D LMMs.} Our comparisons include state-of-the-art 3D LMMs such as 3D-LLM~\cite{3dllm}, LL3DA~\cite{ll3da}, Chat-3D v2~\cite{chat3dv2}, Scene-LLM~\cite{scenellm}, LEO~\cite{leo}, Chat-Scene~\cite{chatscene}, LLaVA-3D~\cite{llava3d}, and Video-3D-LLM~\cite{video3dllm}.

\subsubsection{Metrics}
We use different metrics for the five benchmarks. For Scan2Cap~\cite{scan2cap} and ScanQA~\cite{scanqa}, we evaluate metrics include CIDEr~\cite{cider}, BLEU-4~\cite{bleu}, METEOR~\cite{meteor}, and ROUGE~\cite{rouge}, all of which are at the IoU threshold of 0.5. In addition, we evaluate the exact match accuracy (EM) for ScanQA. For SQA3D~\cite{sqa3d}, we evaluate the EM as well. For ScanRefer~\cite{scanrefer}, we use accuracy at IoU thresholds 0.25 and 0.5. For Multi3DRefer~\cite{multi3drefer}, we compute F1 scores at IoU thresholds 0.25 and 0.5.

\subsubsection{Implementation Details}
\OURS{} is built upon LLaVA-Video-7B~\cite{llavavideo}. We follow the design in Video-3D-LLM~\cite{video3dllm} that takes multi-view renderings from 3D scenes~\cite{scannet} as an input video to the base model, and replace the original learnable 1D positional embeddings with 3D coordinate embeddings. \OURS{} is fine-tuned for one epoch using a combined training set consisting of five datasets: Scan2Cap~\cite{scan2cap}, ScanQA~\cite{scanqa}, SQA3D~\cite{sqa3d}, ScanRefer~\cite{scanrefer}, and Multi3DRefer~\cite{multi3drefer}. 

For training, all videos are downsampled to 3 FPS, with each frame's corresponding camera intrinsics and extrinsics. The input frame resolution is set to $384 \times 384$. We use the AdamW optimizer with a batch size of 8. The warm-up phase spans the first $3\%$ of training steps, during which the learning rate is gradually increased, peaking at 1e-5 for the language model and 2e-6 for the vision encoder. All experiments are conducted on 8 A100 GPUs.

\subsection{Comparisons with State-of-the-arts}

\noindent \textbf{Task 1: 3D Visual Grounding.} In \cref{tab:refertask}, we show results on ScanRefer and Multi3DRefer. \OURS{} achieves state-of-the-art performance across all metrics, surpassing Video-3D-LLM by $+3.9$ Acc@0.25 and $+3.7$ Acc@0.5 on ScanRefer, and by $+2.2$ F1@0.25 and $+2.0$ F1@0.5 on Multi3DRefer.

\begin{table}[t]
\centering
\setlength{\tabcolsep}{5pt}
\resizebox{\linewidth}{!}{
\begin{tabular}{@{}lcccc@{}}
    \toprule
     & \multicolumn{2}{c}{ScanRefer} & \multicolumn{2}{c}{Multi3DRefer}\\
     \cmidrule(lr){2-3} \cmidrule(lr{0pt}){4-5}
     & Acc@0.25 & Acc@0.5 & F1@0.25 & F1@0.5\\
    \midrule
    \multicolumn{1}{@{}l}{\small\textbf{\textit{Expert models}}} \\
    ScanRefer~\cite{scanrefer}	& 37.3	& 24.3 & $-$ & $-$  \\
    MVT~\cite{mvt}	&40.8	&33.3 & $-$ & $-$ \\
    ViL3DRel~\cite{vil3drel} & 47.9	& 37.7 & $-$ & $-$ \\  
    BUTD-DETR~\cite{butddetr}	& 52.2 & 39.8 & $-$ & $-$ \\  
    3DVG-Trans~\cite{3dvgtrans} & 45.9	& 34.5 & $-$ & 25.5 \\
    3DJCG~\cite{3djcg} & 49.6 & 37.3 & $-$ & 26.6 \\
    M3DRef-CLIP~\cite{multi3drefer}	& 51.0 & 44.7	 &42.8 &38.4\\ 
    \midrule
    \multicolumn{1}{@{}l}{\small\textit{\textbf{3D LMMs}}} \\
    3D-LLM~\cite{3dllm}  & 30.3	& $-$ & $-$ & $-$ \\
    Chat-3D v2~\cite{chat3dv2}	& 35.9	& 30.4 & $-$ & $-$ \\
    Grounded 3D-LLM~\cite{grounded3dllm}   & 47.9  & 44.1 & 45.2 & 40.6 \\ 
    Chat-Scene~\cite{chatscene}   & 55.5  & 50.2 & 57.1 & 52.4 \\ 
    LLaVA-3D~\cite{llava3d}	& 50.1 & 42.7 & 49.8 & 43.6 \\
    Video-3D-LLM~\cite{video3dllm} & 58.1 & 51.7 & 58.0 & 52.7 \\
    \midrule
    \textbf{\OURS{} (Ours)} & \textbf{62.0} & \textbf{55.4} & \textbf{60.2} & \textbf{54.7} \\
    \bottomrule
\end{tabular}
}
\caption{Quantitative comparison with SOTA models for 3D Visual Grounding task on ScanRefer and Multi3DRefer.}
\label{tab:refertask}
\end{table}

\noindent \textbf{Task 2: 3D Question Answering.} We report the results on 3D question answering benchmarks (ScanQA and SQA3D) in \cref{tab:scanqa_sqa3d}. \OURS{} consistently outperforms the previous state-of-the-art Video-3D-LLM, with improvements of approximately 5–10$\%$ across all metrics.

\begin{table}[ht]
\centering
\resizebox{\linewidth}{!}{
\setlength{\tabcolsep}{2pt}
\begin{tabular}{@{}lcccccccccc@{}}
    \toprule
     & \multicolumn{5}{c}{ScanQA (val)} & SQA3D (test) \\
     \cmidrule(l{1pt}r{1pt}){2-6} \cmidrule(l{1pt}r{0pt}){7-7}
      & CIDEr & BLEU-4 & METEOR & ROUGE & EM & EM \\
    \midrule
    \multicolumn{1}{@{}l}{\small\textbf{\textit{Expert models}}} \\
    Scan2Cap~\cite{scan2cap} & $-$ & $-$ & $-$ & $-$ & $-$ & 41.0 \\
    ClipBERT~\cite{clipbert} & $-$ & $-$ & $-$ & $-$ & $-$ & 43.3 \\
    ScanRefer+MCAN~\cite{mcan} & 55.4 & 7.9 & 11.5 & 30.0 & 18.6 & - \\
    ScanQA~\cite{scanqa} & 64.9 & 10.1 & 13.1 & 33.3 & 21.1 & 47.2 \\
    3D-VisTA~\cite{3dvista} & 69.6 & 10.4 & 13.9 & 35.7 & 22.4 & 48.5 \\
    \midrule
    \multicolumn{1}{@{}l}{\small\textit{\textbf{Zero-shot 2D LMMs}}} \\
    VideoChat2~\cite{videochat2} & 49.2 & 9.6 & 9.5 & 28.2 & 19.2 & 37.3 \\
    Qwen2.5-VL-7B~\cite{qwen25} & 53.9 & 3.0 & 11.4 & 29.3 & $-$ & 46.5 \\
    LLaVA-Video~\cite{llavavideo} & 88.7 & 3.1 & 17.7 & 44.6 & $-$ & 48.5 \\
    \midrule
    \multicolumn{1}{@{}l}{\small\textit{\textbf{3D LMMs}}} \\
    3D-LLM~\cite{3dllm} & 69.4 & 12.0 & 14.5 & 35.7 & 20.5 & $-$ \\
    LL3DA~\cite{ll3da} & 76.8 & 13.5 & 15.9 & 37.3 & $-$ & $-$ \\
    Chat-3D v2~\cite{chat3dv2} & 87.6 & 14.0 & $-$ & $-$ & $-$ & 54.7\\
    Scene-LLM ~\cite{scenellm} & 80.0 & 12.0 & 16.6 & 40.0 &27.2  & 54.2\\
    LEO~\cite{leo} &101.4 & 13.2 & 20.0 & 49.2 & 24.5 & 50.0 \\
    Chat-Scene ~\cite{chatscene} & 87.7 & 14.3 & 18.0 & 41.6 & 21.6  & 54.6 \\
    LLaVA-3D~\cite{llava3d} & 91.7 & 14.5 & 20.7 & 50.1 & 27.0 & 55.6 \\
    Video-3D-LLM~\cite{video3dllm} & 102.1 & 16.2 & 19.8 & 49.0 & 30.1  & 58.6 \\
    \midrule
    \textbf{\OURS{} (Ours)} & \textbf{107.1} & \textbf{17.8} & \textbf{20.8} & \textbf{50.9} & \textbf{31.2} & \textbf{62.3} \\
    \bottomrule
\end{tabular}
}
\caption{Quantitative comparisons with SOTA models for 3D Question Answering task on ScanQA and SQA3D.}
\label{tab:scanqa_sqa3d}
\end{table}

\begin{table}[ht]
    \label{tab:benchmark-3d-dense-cap}
    \centering
    \setlength{\tabcolsep}{10pt}
    \resizebox{\linewidth}{!}{
    \begin{tabular}{@{}lcccc@{}}
    \toprule
    & \multicolumn{4}{c}{Scan2Cap (Val)} \\ 
    \cmidrule(lr{0pt}){2-5}  & CIDEr  & BLEU-4 & METEOR & ROUGE \\
    \midrule
    \multicolumn{1}{@{}l}{\small\textbf{\textit{Expert models}}} \\
    Scan2Cap~\cite{scan2cap}                & 39.1           & 23.3             & 22.0           & 44.8      \\
    3DJCG~\cite{3djcg}  & 49.5           & 31.0           & 24.2           & 50.8       \\
    3D-VLP~\cite{3dvlp}              & 55.0           & 32.3             & 24.8           & 51.5          \\
    3D-VisTA~\cite{3dvista}        & 61.6          & 34.1           & 26.8         & 55.0     \\
    Vote2Cap-DETR~\cite{vote2cap}       & 61.8           & 34.5            & 26.2          & 54.4         \\
    \midrule
    \multicolumn{1}{@{}l}{\small\textbf{\textit{3D LMMs}}} \\
    LL3DA~\cite{ll3da}          & 65.2            & 36.8      & 26.0          & 55.0                  \\
    LEO~\cite{leo}    &68.4 &36.9 &27.7 &57.8 \\
    Chat-Scene~\cite{chatscene} &77.2 &36.3 &28.0 & 58.1 \\
    LLaVA-3D~\cite{llava3d}  & 79.2 & 41.1 & \textbf{30.2}  & \textbf{63.4} \\
    Video-3D-LLM~\cite{video3dllm} & 83.8 & 41.3 & 28.9 & 62.3 \\
    \midrule
    \textbf{\OURS{} (Ours)} & \textbf{90.5} & \textbf{41.7} & 28.9 & 62.2 \\
    \bottomrule
    \end{tabular}
    }
    \caption{Quantitative comparisons with SOTA models for 3D Dense Captioning task on Scan2Cap.}
    \label{tab:scan2cap}
\end{table}

\noindent \textbf{Task 3: 3D Dense Captioning.} \cref{tab:scan2cap} presents the performance of \OURS{} and baseline methods on Scan2Cap. \OURS{} achieves results comparable to prior state-of-the-art models such as LLaVA-3D and Video-3D-LLM in terms of METEOR and ROUGE, while surpassing them on CIDEr ($+6.7$) and BLEU-4 ($+0.4$).

Overall, \OURS{} consistently outperforms existing methods across tasks and metrics, demonstrating the effectiveness of our proposed designs.

\begin{table*}[ht]{
\centering
\small
\resizebox{\linewidth}{!}{
\setlength{\tabcolsep}{4pt}
\begin{tabular}{lccccccccc}
\toprule 
\multirow{2}{*}[-2pt]{Ablation Setting} & \multicolumn{2}{c}{Scan2Cap}  & \multicolumn{2}{c}{ScanQA} & \multicolumn{1}{c}{SQA3D}  & \multicolumn{2}{c}{ScanRefer} & \multicolumn{2}{c}{Multi3DRef} \\ 
\cmidrule(lr){2-3} \cmidrule(lr){4-5} \cmidrule(lr){6-6} \cmidrule(lr){7-8} \cmidrule(lr){9-10} 
 & BLEU-4 & CIDEr & CIDEr & EM & EM & Acc@0.25 & Acc@0.5 & F1@0.25 & F1@0.5\\
 \midrule
 \OURS{}  & 41.7 & 90.5 & 107.1 & 31.2 & 62.3 & 62.0 & 55.4 & 60.2 & 54.7 \\
 \midrule
 \multicolumn{5}{l}{\textit{\textbf{a) without...}}}\\
 \Grid{} Grid & 41.3 & 84.4 & 103.9 & 30.0 & 59.2 & 62.0 & 55.1 & 59.9 & 54.6 \\
  \Token{} Token & 40.8 & 85.8 & 107.6 & 31.5 & 62.2 & 59.5 & 53.0 & 58.7 & 53.3 \\

\midrule
\multicolumn{5}{l}{\textit{\textbf{b) \Grid{} Grid setting}}}\\
Grid Size = 10 & 41.4 & 85.8 & 104.8 & 30.3 & 59.6 & 61.5 & 55.0 & 60.1 & 54.7 \\
Grid Size = 16 & 42.2 & 88.8 & 102.8 & 29.7 & 58.8 & 61.7 & 55.1 & 60.0 & 54.5 \\
Grid Size = 24 & 42.1 & 87.3 & 102.7 & 30.0 & 58.8 & 61.9 & 55.3 & 59.7 & 54.2 \\
Replace texts w/ embeddings & 41.9 & 86.1 & 103.6 & 30.0 & 59.1 & 61.8 & 55.3 & 59.9 & 54.6 \\
\midrule
\multicolumn{5}{l}{\textit{\textbf{c) \Token{} Token setting}}}\\
Contrastive $\xrightarrow{}$ Regression & 40.8 & 86.3 & 106.7 & 31.5 & 59.8 & 59.0 & 52.7 & 59.1 & 53.7 \\
GT Boxes $\xrightarrow{}$ All Related Boxes & 40.3 & 79.6 & 98.7 & 28.4 & 57.1 & 57.2 & 51.0 & 56.3 & 51.3 \\

\bottomrule
\end{tabular}
}
\caption{Ablation studies on \Grid{} Grid and \Token{} Token and their respective settings.} 
\label{tab:ablation}
}
\end{table*}

\subsection{Ablation Study}
We provide ablation studies on the proposed \Grid{} Grid and \Token{} Token in \cref{tab:ablation}.

\noindent \textbf{\Grid{} Grid and \Token{} Token.} Section a) of \cref{tab:ablation} show the ablation results when each of the two proposed modules is removed individually. Performance drops in both cases, confirming the contribution of each component. The \token{} token provides a more significant boost in visual grounding tasks by improving model's object localization capacity. In contrast, the \grid{} grid offers greater benefits for dense captioning and question answering by supplying high-level abstractions of the 3D environment, enhancing overall scene understanding.

\noindent \textbf{\Grid{} Grid Size.}
We ablate different settings of \grid{} grid in Section b) of \cref{tab:ablation}. We first expanding the default $6\times6$ grid to $10\times10$, $16\times16$, and $24\times24$ layouts. Finer grids could minimize the chance of multiple objects sharing a single cell, and more precisely represent the layout of 3D scenes. However, the results show no performance gains from higher-resolution grids. We suspect that coarser, more abstract grids provide clearer structural cues, while overly detailed grids introduce unnecessary complexity that hinders scene understanding for the model. 

\noindent \textbf{\Grid{} Grid Object Representation.}
By default, we represent \grid{} grids with text, directly incorporating object names as textual elements. To examine whether an “object-centric” design, as in prior works~\cite{chat3dv2, leo, chatscene}, improves performance, we replace object names with embeddings extracted from the input video using a pretrained vision encoder, followed by pooling, identical to the embeddings used for \token{} tokens. As shown by the last row of Section b) in \cref{tab:ablation}, this embedding-based variant underperforms our text-only design. We attribute this to modality mismatch: vision-derived embeddings introduce alignment errors, which are further amplified by 3D aggregation, whereas text tokens are compatible with the base model.

\noindent \textbf{\Token{} Token Settings.}
In Section c) of Tab.~\ref{tab:ablation}, We ablate two alternative designs for our \token{} token. We first replace the InfoNCE contrastive loss with a mean squared error (MSE) loss for direct regression, and the experiment show that this modification yields poorer performance. We also experiment with using all related boxes for supervision instead of just the ground‑truth box. Concretely, we parse each training sample’s questions and answers to identify every referenced object, then label every bounding box matching those names in the 3D scene as relevant. However, this modification results in a significant decline in performance. We attribute this to the increased ambiguity introduced by multiple instances of the same object category, which can confuse the model and degrade the precision of attention.

\begin{figure}[b]
    \centering
    \includegraphics[width=\linewidth,trim=0 0 0 1em,clip]{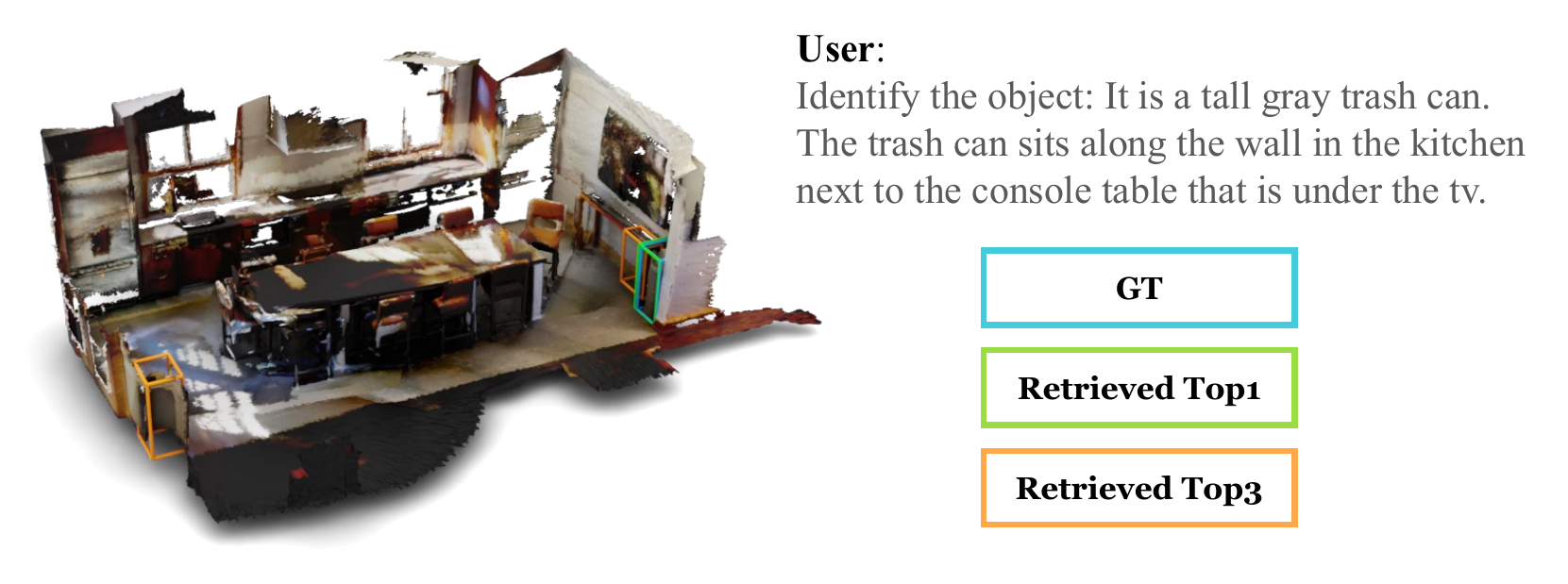}
    \caption{Object retrieval results using our \token{} token. We compute similarities between the output \token{} embedding and all object embeddings in the scene. The top-1 result is shown in a green box; top-3 results are highlighted in orange.}
    \vspace{-1mm}
    \label{fig:retrieval}
\end{figure}

\begin{figure*}[t]
    \centering
    \includegraphics[width=\linewidth]{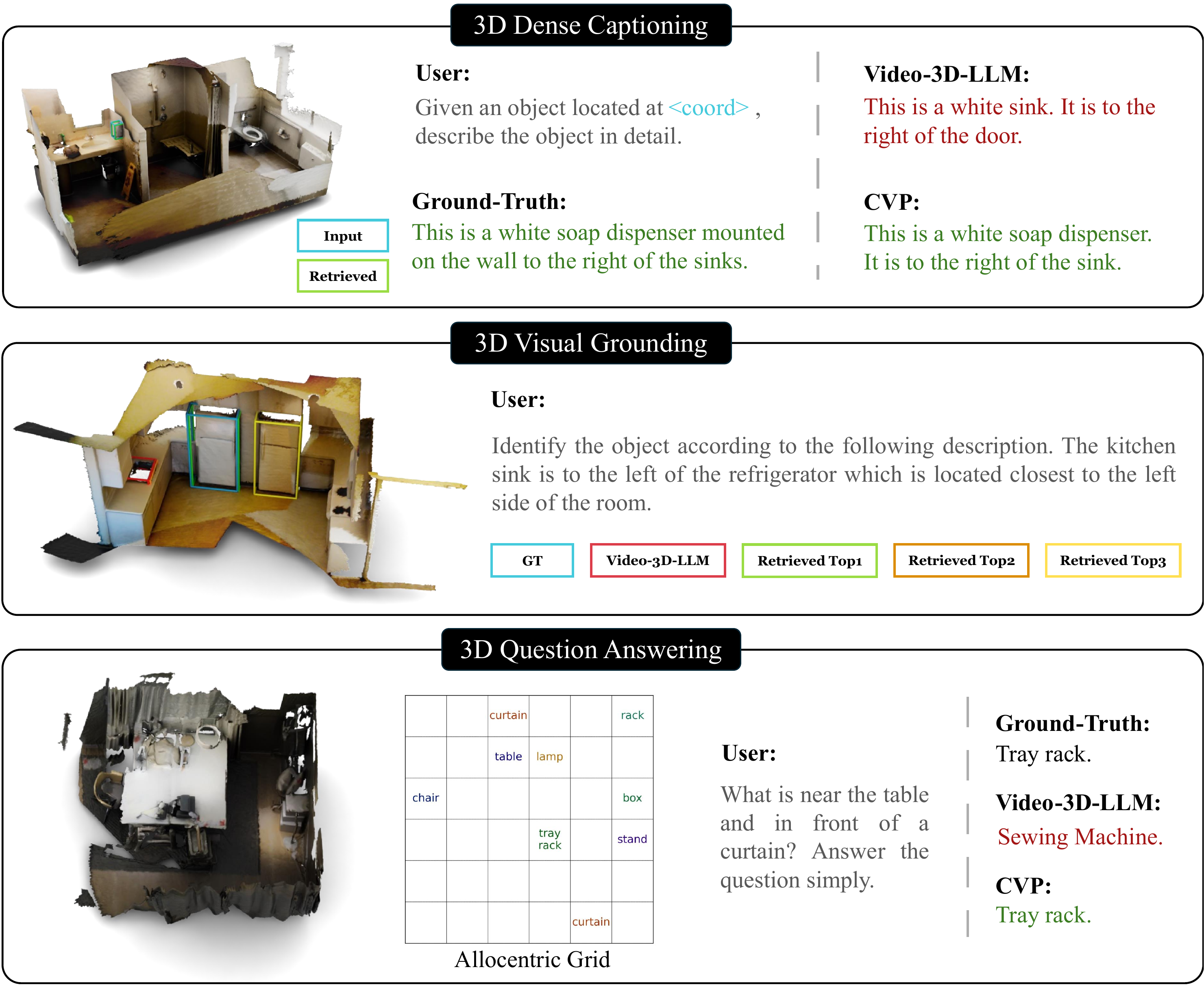}
    \caption{Qualitative comparisons across three representative 3D vision-language tasks: dense captioning, visual grounding, and question answering. \OURS{} consistently outperforms the baseline model.}
    \label{fig:scan2cap}
    \vspace{-1em}
\end{figure*}

\subsection{Visualizations}
We showcase the effectiveness of the \token{} token through an object retrieval task. Specifically, we compute embeddings for all objects in the scene (as described earlier), and retrieve those most similar to the output \token{} embedding. Representative results are shown in \cref{fig:retrieval}. The top-1 retrieved object consistently matches the one referred to in the user query. The top-3 results often include semantically related objects (\eg, a trash can at the bottom left) and spatially adjacent instances (\eg, a nearby cabinet), illustrating that the \token{} token embedding effectively captures information relevant to the current question and its potential answers.

We also present visualization of the three 3D spatial understanding tasks in \cref{fig:scan2cap}:

\begin{itemize}
    \item \textbf{3D Dense Captioning.} Given the object location, Video-3D-LLM mistakes a nearby white sink for the target soap dispenser. In contrast, \OURS{} accurately describes the soap dispenser. The improvement comes from the \token{} token, which helps focus on the correct object. 

    \item \textbf{3D Visual Grounding.} When asked to identify the kitchen sink relative to the refrigerator, Video-3D-LLM selects the wrong region. \OURS{} successfully grounds the correct object. It leverages retrieved candidates and spatial language cues to resolve the relational instruction. 

    \item \textbf{3D Question Answering.} The model is queried about an object near the table and in front of a curtain. Video-3D-LLM responds with ``sewing machine'', which violates the spatial constraint. \OURS{} outputs ``tray rack'' by reasoning over the \grid{} grid. 
\end{itemize}

\section{Conclusion}

We present \OURS{}, a 3D scene understanding framework inspired by components of the human visual field. \OURS{} introduces two new key mechanisms: the \token{} token that guides the model’s attention toward target objects, analogous to central vision for fine-grained perception; and the \grid{} grid, which captures global scene context and spatial relationships, akin to peripheral vision. These complementary components enable structured, context-aware reasoning in complex 3D environments. Extensive experiments across multiple benchmarks demonstrate that \OURS{} achieves state-of-the-art performance.

\section*{Acknowledgment} This work is supported by NSF award IIS-2127544 and NSF award IIS-2433768.
We thank Lambda, Inc. for their compute resource help.

{
    \small
    \bibliographystyle{ieeenat_fullname}
    \bibliography{main}

@String(CVPR= {IEEE Conf. Comput. Vis. Pattern Recog.})

@String(ICCV= {Int. Conf. Comput. Vis.})

@String(ECCV= {Eur. Conf. Comput. Vis.})

@String(ICLR = {Int. Conf. Learn. Represent.})

@String(CVPR  = {CVPR})

@String(ICCV  = {ICCV})

@String(ECCV  = {ECCV})

@String(ICLR  = {ICLR})

@inproceedings{scanrefer,
  title={Scanrefer: 3d object localization in rgb-d scans using natural language},
  author={Chen, Dave Zhenyu and Chang, Angel X and Nie{\ss}ner, Matthias},
  booktitle={ECCV},
  pages={202--221},
  year={2020},
  organization={Springer}
}

@inproceedings{multi3drefer,
  title={Multi3drefer: Grounding text description to multiple 3d objects},
  author={Zhang, Yiming and Gong, ZeMing and Chang, Angel X},
  booktitle={ICCV},
  pages={15225--15236},
  year={2023}
}

@inproceedings{sqa3d,
  title={SQA3D: Situated Question Answering in 3D Scenes},
  author={Ma, Xiaojian and Yong, Silong and Zheng, Zilong and Li, Qing and Liang, Yitao and Zhu, Song-Chun and Huang, Siyuan},
  booktitle={ICLR},
  year={2023},
}

@inproceedings{scan2cap,
  title={Scan2cap: Context-aware dense captioning in rgb-d scans},
  author={Chen, Zhenyu and Gholami, Ali and Nie{\ss}ner, Matthias and Chang, Angel X},
  booktitle={CVPR},
  pages={3193--3203},
  year={2021}
}

@inproceedings{scanqa,
  title={Scanqa: 3d question answering for spatial scene understanding},
  author={Azuma, Daichi and Miyanishi, Taiki and Kurita, Shuhei and Kawanabe, Motoaki},
  booktitle={CVPR},
  pages={19129--19139},
  year={2022}
}

@inproceedings{video3dllm,
  title={Video-3d llm: Learning position-aware video representation for 3d scene understanding},
  author={Zheng, Duo and Huang, Shijia and Wang, Liwei},
  booktitle={CVPR},
  pages={8995--9006},
  year={2025}
}

@inproceedings{scannet,
  title={Scannet: Richly-annotated 3d reconstructions of indoor scenes},
  author={Dai, Angela and Chang, Angel X and Savva, Manolis and Halber, Maciej and Funkhouser, Thomas and Nie{\ss}ner, Matthias},
  booktitle={CVPR},
  pages={5828--5839},
  year={2017}
}

@inproceedings{mcan,
  title={Deep modular co-attention networks for visual question answering},
  author={Yu, Zhou and Yu, Jun and Cui, Yuhao and Tao, Dacheng and Tian, Qi},
  booktitle={CVPR},
  pages={6281--6290},
  year={2019}
}

@inproceedings{clipbert,
  title={Less is more: Clipbert for video-and-language learning via sparse sampling},
  author={Lei, Jie and Li, Linjie and Zhou, Luowei and Gan, Zhe and Berg, Tamara L and Bansal, Mohit and Liu, Jingjing},
  booktitle={CVPR},
  pages={7331--7341},
  year={2021}
}

@inproceedings{3dvista,
  title={3d-vista: Pre-trained transformer for 3d vision and text alignment},
  author={Zhu, Ziyu and Ma, Xiaojian and Chen, Yixin and Deng, Zhidong and Huang, Siyuan and Li, Qing},
  booktitle={ICCV},
  pages={2911--2921},
  year={2023}
}

@article{3dllm,
  title={3d-llm: Injecting the 3d world into large language models},
  author={Hong, Yining and Zhen, Haoyu and Chen, Peihao and Zheng, Shuhong and Du, Yilun and Chen, Zhenfang and Gan, Chuang},
  journal={NeurIPS},
  volume={36},
  pages={20482--20494},
  year={2023}
}

@inproceedings{ll3da,
  title={Ll3da: Visual interactive instruction tuning for omni-3d understanding reasoning and planning},
  author={Chen, Sijin and Chen, Xin and Zhang, Chi and Li, Mingsheng and Yu, Gang and Fei, Hao and Zhu, Hongyuan and Fan, Jiayuan and Chen, Tao},
  booktitle={CVPR},
  pages={26428--26438},
  year={2024}
}

@article{chat3dv2,
  title={Chat-3d v2: Bridging 3d scene and large language models with object identifiers},
  author={Huang, Haifeng and Wang, Zehan and Huang, Rongjie and Liu, Luping and Cheng, Xize and Zhao, Yang and Jin, Tao and Zhao, Zhou},
  journal={arXiv preprint arXiv:2312.08168},
  year={2023}
}

@article{chatscene,
  title={Chat-scene: Bridging 3d scene and large language models with object identifiers},
  author={Huang, Haifeng and Chen, Yilun and Wang, Zehan and Huang, Rongjie and Xu, Runsen and Wang, Tai and Liu, Luping and Cheng, Xize and Zhao, Yang and Pang, Jiangmiao and others},
  journal={NeurIPS},
  volume={37},
  pages={113991--114017},
  year={2024}
}

@article{scenellm,
  title={Scene-llm: Extending language model for 3d visual understanding and reasoning},
  author={Fu, Rao and Liu, Jingyu and Chen, Xilun and Nie, Yixin and Xiong, Wenhan},
  journal={arXiv preprint arXiv:2403.11401},
  year={2024}
}

@inproceedings{leo,
  title={An embodied generalist agent in 3D world},
  author={Huang, Jiangyong and Yong, Silong and Ma, Xiaojian and Linghu, Xiongkun and Li, Puhao and Wang, Yan and Li, Qing and Zhu, Song-Chun and Jia, Baoxiong and Huang, Siyuan},
  booktitle={ICML},
  pages={20413--20451},
  year={2024}
}

@article{llavavideo,
  title={Video instruction tuning with synthetic data},
  author={Zhang, Yuanhan and Wu, Jinming and Li, Wei and Li, Bo and Ma, Zejun and Liu, Ziwei and Li, Chunyuan},
  journal={arXiv preprint arXiv:2410.02713},
  year={2024}
}

@inproceedings{llava3d,
  title={Llava-3d: A simple yet effective pathway to empowering lmms with 3d capabilities},
  author={Zhu, Chenming and Wang, Tai and Zhang, Wenwei and Pang, Jiangmiao and Liu, Xihui},
  booktitle={ICCV},
  pages={4295--4305},
  year={2025}
}

@inproceedings{3dvlp,
  title={Context-aware alignment and mutual masking for 3d-language pre-training},
  author={Jin, Zhao and Hayat, Munawar and Yang, Yuwei and Guo, Yulan and Lei, Yinjie},
  booktitle={CVPR},
  pages={10984--10994},
  year={2023}
}

@inproceedings{mvt,
  title={Multi-view transformer for 3d visual grounding},
  author={Huang, Shijia and Chen, Yilun and Jia, Jiaya and Wang, Liwei},
  booktitle={CVPR},
  pages={15524--15533},
  year={2022}
}

@article{dahnert2021panoptic,
  title={Panoptic 3d scene reconstruction from a single rgb image},
  author={Dahnert, Manuel and Hou, Ji and Nie{\ss}ner, Matthias and Dai, Angela},
  journal={NeurIPS},
  volume={34},
  pages={8282--8293},
  year={2021}
}

@inproceedings{zhang2023uni,
  title={Uni-3d: A universal model for panoptic 3d scene reconstruction},
  author={Zhang, Xiang and Chen, Zeyuan and Wei, Fangyin and Tu, Zhuowen},
  booktitle={ICCV},
  pages={9256--9266},
  year={2023}
}

@inproceedings{zhao2025depr,
  title={Depr: Depth guided single-view scene reconstruction with instance-level diffusion},
  author={Zhao, Qingcheng and Zhang, Xiang and Xu, Haiyang and Chen, Zeyuan and Xie, Jianwen and Gao, Yuan and Tu, Zhuowen},
  booktitle={ICCV},
  pages={5722--5733},
  year={2025}
}

@inproceedings{videochat2,
  title={Mvbench: A comprehensive multi-modal video understanding benchmark},
  author={Li, Kunchang and Wang, Yali and He, Yinan and Li, Yizhuo and Wang, Yi and Liu, Yi and Wang, Zun and Xu, Jilan and Chen, Guo and Luo, Ping and others},
  booktitle={CVPR},
  pages={22195--22206},
  year={2024}
}

@inproceedings{3djcg,
  title={3djcg: A unified framework for joint dense captioning and visual grounding on 3d point clouds},
  author={Cai, Daigang and Zhao, Lichen and Zhang, Jing and Sheng, Lu and Xu, Dong},
  booktitle={CVPR},
  pages={16464--16473},
  year={2022}
}

@inproceedings{3dvgtrans,
  title={3dvg-transformer: Relation modeling for visual grounding on point clouds},
  author={Zhao, Lichen and Cai, Daigang and Sheng, Lu and Xu, Dong},
  booktitle={ICCV},
  pages={2928--2937},
  year={2021}
}

@inproceedings{butddetr,
  title={Bottom up top down detection transformers for language grounding in images and point clouds},
  author={Jain, Ayush and Gkanatsios, Nikolaos and Mediratta, Ishita and Fragkiadaki, Katerina},
  booktitle={ECCV},
  pages={417--433},
  year={2022},
  organization={Springer}
}

@article{grounded3dllm,
  title={Grounded 3d-llm with referent tokens},
  author={Chen, Yilun and Yang, Shuai and Huang, Haifeng and Wang, Tai and Xu, Runsen and Lyu, Ruiyuan and Lin, Dahua and Pang, Jiangmiao},
  journal={arXiv preprint arXiv:2405.10370},
  year={2024}
}

@article{vil3drel,
  title={Language conditioned spatial relation reasoning for 3d object grounding},
  author={Chen, Shizhe and Guhur, Pierre-Louis and Tapaswi, Makarand and Schmid, Cordelia and Laptev, Ivan},
  journal={NeurIPS},
  volume={35},
  pages={20522--20535},
  year={2022}
}

@inproceedings{vote2cap,
  title={End-to-end 3d dense captioning with vote2cap-detr},
  author={Chen, Sijin and Zhu, Hongyuan and Chen, Xin and Lei, Yinjie and Yu, Gang and Chen, Tao},
  booktitle={CVPR},
  pages={11124--11133},
  year={2023}
}

@article{
  llavaonevision,
  title={{LL}a{VA}-OneVision: Easy Visual Task Transfer},
  author={Bo Li and Yuanhan Zhang and Dong Guo and Renrui Zhang and Feng Li and Hao Zhang and Kaichen Zhang and Peiyuan Zhang and Yanwei Li and Ziwei Liu and Chunyuan Li},
  journal={TMLR},
  issn={2835-8856},
  year={2025},
  url={https://openreview.net/forum?id=zKv8qULV6n},
  note={}
}

@article{qwen25,
  title={Qwen2. 5-vl technical report},
  author={Bai, Shuai and Chen, Keqin and Liu, Xuejing and Wang, Jialin and Ge, Wenbin and Song, Sibo and Dang, Kai and Wang, Peng and Wang, Shijie and Tang, Jun and others},
  journal={arXiv preprint arXiv:2502.13923},
  year={2025}
}

@inproceedings{pointllm,
  title={Pointllm: Empowering large language models to understand point clouds},
  author={Xu, Runsen and Wang, Xiaolong and Wang, Tai and Chen, Yilun and Pang, Jiangmiao and Lin, Dahua},
  booktitle={ECCV},
  pages={131--147},
  year={2024},
  organization={Springer}
}

@inproceedings{pointbert,
  title={Point-bert: Pre-training 3d point cloud transformers with masked point modeling},
  author={Yu, Xumin and Tang, Lulu and Rao, Yongming and Huang, Tiejun and Zhou, Jie and Lu, Jiwen},
  booktitle={CVPR},
  pages={19313--19322},
  year={2022}
}

@inproceedings{clip,
  title={Learning transferable visual models from natural language supervision},
  author={Radford, Alec and Kim, Jong Wook and Hallacy, Chris and Ramesh, Aditya and Goh, Gabriel and Agarwal, Sandhini and Sastry, Girish and Askell, Amanda and Mishkin, Pamela and Clark, Jack and others},
  booktitle={ICML},
  year={2021},
}

@inproceedings{voxelnet,
  title={Voxelnet: End-to-end learning for point cloud based 3d object detection},
  author={Zhou, Yin and Tuzel, Oncel},
  booktitle={CVPR},
  pages={4490--4499},
  year={2018}
}

@inproceedings{pointpillars,
  title={Pointpillars: Fast encoders for object detection from point clouds},
  author={Lang, Alex H and Vora, Sourabh and Caesar, Holger and Zhou, Lubing and Yang, Jiong and Beijbom, Oscar},
  booktitle={CVPR},
  pages={12697--12705},
  year={2019}
}

@inproceedings{frustumpointnet,
  title={Frustum pointnets for 3d object detection from rgb-d data},
  author={Qi, Charles R and Liu, Wei and Wu, Chenxia and Su, Hao and Guibas, Leonidas J},
  booktitle={CVPR},
  pages={918--927},
  year={2018}
}

@article{thrun2002probabilistic,
  title={Probabilistic robotics},
  author={Thrun, Sebastian},
  journal={Communications of the ACM},
  volume={45},
  number={3},
  pages={52--57},
  year={2002},
  publisher={ACM New York, NY, USA}
}

@inproceedings{salas2013slam++,
  title={Slam++: Simultaneous localisation and mapping at the level of objects},
  author={Salas-Moreno, Renato F and Newcombe, Richard A and Strasdat, Hauke and Kelly, Paul HJ and Davison, Andrew J},
  booktitle={CVPR},
  pages={1352--1359},
  year={2013}
}

@inproceedings{newcombe2011kinectfusion,
  title={Kinectfusion: Real-time dense surface mapping and tracking},
  author={Newcombe, Richard A and Izadi, Shahram and Hilliges, Otmar and Molyneaux, David and Kim, David and Davison, Andrew J and Kohi, Pushmeet and Shotton, Jamie and Hodges, Steve and Fitzgibbon, Andrew},
  booktitle={IEEE international symposium on mixed and augmented reality},
  pages={127--136},
  year={2011}
}

@inproceedings{pointclip,
  title={Pointclip: Point cloud understanding by clip},
  author={Zhang, Renrui and Guo, Ziyu and Zhang, Wei and Li, Kunchang and Miao, Xupeng and Cui, Bin and Qiao, Yu and Gao, Peng and Li, Hongsheng},
  booktitle={CVPR},
  pages={8552--8562},
  year={2022}
}

@inproceedings{more,
  title={More: Multi-order relation mining for dense captioning in 3d scenes},
  author={Jiao, Yang and Chen, Shaoxiang and Jie, Zequn and Chen, Jingjing and Ma, Lin and Jiang, Yu-Gang},
  booktitle={ECCV},
  pages={528--545},
  year={2022},
  organization={Springer}
}

@inproceedings{xtrans2cap,
  title={X-trans2cap: Cross-modal knowledge transfer using transformer for 3d dense captioning},
  author={Yuan, Zhihao and Yan, Xu and Liao, Yinghong and Guo, Yao and Li, Guanbin and Cui, Shuguang and Li, Zhen},
  booktitle={CVPR},
  pages={8563--8573},
  year={2022}
}

@inproceedings{parelli2023clip,
  title={Clip-guided vision-language pre-training for question answering in 3d scenes},
  author={Parelli, Maria and Delitzas, Alexandros and Hars, Nikolas and Vlassis, Georgios and Anagnostidis, Sotirios and Bachmann, Gregor and Hofmann, Thomas},
  booktitle={CVPR},
  pages={5607--5612},
  year={2023}
}

@inproceedings{multiviewground,
  title={Multi-view transformer for 3d visual grounding},
  author={Huang, Shijia and Chen, Yilun and Jia, Jiaya and Wang, Liwei},
  booktitle={CVPR},
  pages={15524--15533},
  year={2022}
}

@inproceedings{3drpnet,
  title={3DRP-Net: 3D Relative Position-aware Network for 3D Visual Grounding},
  author={Wang, Zehan and Huang, Haifeng and Zhao, Yang and Li, Linjun and Cheng, Xize and Zhu, Yichen and Yin, Aoxiong and Zhao, Zhou},
  booktitle={EMNLP},
  year={2023}
}

@inproceedings{qformer,
  title={Blip-2: Bootstrapping language-image pre-training with frozen image encoders and large language models},
  author={Li, Junnan and Li, Dongxu and Savarese, Silvio and Hoi, Steven},
  booktitle={ICML},
  year={2023},
}

@inproceedings{bleu,
  title={Bleu: a method for automatic evaluation of machine translation},
  author={Papineni, Kishore and Roukos, Salim and Ward, Todd and Zhu, Wei-Jing},
  booktitle={ACL},
  year={2002}
}

@inproceedings{meteor,
  title={METEOR: An automatic metric for MT evaluation with improved correlation with human judgments},
  author={Banerjee, Satanjeev and Lavie, Alon},
  booktitle={ACL workshop on intrinsic and extrinsic evaluation measures for machine translation and/or summarization},
  year={2005}
}

@inproceedings{cider,
  title={Cider: Consensus-based image description evaluation},
  author={Vedantam, Ramakrishna and Lawrence Zitnick, C and Parikh, Devi},
  booktitle={CVPR},
  pages={4566--4575},
  year={2015}
}

@inproceedings{rouge,
  title={Rouge: A package for automatic evaluation of summaries},
  author={Lin, Chin-Yew},
  booktitle={Text summarization branches out},
  year={2004}
}

@inproceedings{zhai2023sigmoid,
  title={Sigmoid loss for language image pre-training},
  author={Zhai, Xiaohua and Mustafa, Basil and Kolesnikov, Alexander and Beyer, Lucas},
  booktitle={ICCV},
  pages={11975--11986},
  year={2023}
}

@article{oord2018representation,
  title={Representation learning with contrastive predictive coding},
  author={Oord, Aaron van den and Li, Yazhe and Vinyals, Oriol},
  journal={arXiv preprint arXiv:1807.03748},
  year={2018}
}

@inproceedings{eda,
  title={Eda: Explicit text-decoupling and dense alignment for 3d visual grounding},
  author={Wu, Yanmin and Cheng, Xinhua and Zhang, Renrui and Cheng, Zesen and Zhang, Jian},
  booktitle={CVPR},
  pages={19231--19242},
  year={2023}
}

@inproceedings{achlioptas2020referit3d,
  title={Referit3d: Neural listeners for fine-grained 3d object identification in real-world scenes},
  author={Achlioptas, Panos and Abdelreheem, Ahmed and Xia, Fei and Elhoseiny, Mohamed and Guibas, Leonidas},
  booktitle={ECCV},
  pages={422--440},
  year={2020},
  organization={Springer}
}

@inproceedings{man2024situational,
  title={Situational awareness matters in 3d vision language reasoning},
  author={Man, Yunze and Gui, Liang-Yan and Wang, Yu-Xiong},
  booktitle={CVPR},
  pages={13678--13688},
  year={2024}
}

@article{llava,
  title={Visual instruction tuning},
  author={Liu, Haotian and Li, Chunyuan and Wu, Qingyang and Lee, Yong Jae},
  journal={NeurIPS},
  volume={36},
  pages={34892--34916},
  year={2023}
}

@article{embodied1,
  title={Semantic mapping for mobile robots in indoor scenes: A survey},
  author={Han, Xiaoning and Li, Shuailong and Wang, Xiaohui and Zhou, Weijia},
  journal={Information},
  volume={12},
  number={2},
  pages={92},
  year={2021},
  publisher={MDPI}
}

@inproceedings{embodied2,
  title={Move to understand a 3d scene: Bridging visual grounding and exploration for efficient and versatile embodied navigation},
  author={Zhu, Ziyu and Wang, Xilin and Li, Yixuan and Zhang, Zhuofan and Ma, Xiaojian and Chen, Yixin and Jia, Baoxiong and Liang, Wei and Yu, Qian and Deng, Zhidong and others},
  booktitle={ICCV},
  pages={8120--8132},
  year={2025}
}

@article{embodied3,
  title={Embodied Intelligence for 3D Understanding: A Survey on 3D Scene Question Answering},
  author={Li, Zechuan and Yu, Hongshan and Ding, Yihao and Li, Yan and He, Yong and Akhtar, Naveed},
  journal={arXiv preprint arXiv:2502.00342},
  year={2025}
}

@inproceedings{embodied4,
  title={3D-mem: 3D scene memory for embodied exploration and reasoning},
  author={Yang, Yuncong and Yang, Han and Zhou, Jiachen and Chen, Peihao and Zhang, Hongxin and Du, Yilun and Gan, Chuang},
  booktitle={CVPR},
  pages={17294--17303},
  year={2025}
}

@inproceedings{min2024driveworld,
  title={Driveworld: 4d pre-trained scene understanding via world models for autonomous driving},
  author={Min, Chen and Zhao, Dawei and Xiao, Liang and Zhao, Jian and Xu, Xinli and Zhu, Zheng and Jin, Lei and Li, Jianshu and Guo, Yulan and Xing, Junliang and others},
  booktitle={CVPR},
  pages={15522--15533},
  year={2024}
}

@article{muhammad2022vision,
  title={Vision-based semantic segmentation in scene understanding for autonomous driving: Recent achievements, challenges, and outlooks},
  author={Muhammad, Khan and Hussain, Tanveer and Ullah, Hayat and Del Ser, Javier and Rezaei, Mahdi and Kumar, Neeraj and Hijji, Mohammad and Bellavista, Paolo and De Albuquerque, Victor Hugo C},
  journal={IEEE Transactions on Intelligent Transportation Systems},
  volume={23},
  number={12},
  pages={22694--22715},
  year={2022},
  publisher={IEEE}
}

@article{he2025asap,
  title={Asap: Aligning simulation and real-world physics for learning agile humanoid whole-body skills},
  author={He, Tairan and Gao, Jiawei and Xiao, Wenli and Zhang, Yuanhang and Wang, Zi and Wang, Jiashun and Luo, Zhengyi and He, Guanqi and Sobanbab, Nikhil and Pan, Chaoyi and others},
  journal={arXiv preprint arXiv:2502.01143},
  year={2025}
}

@inproceedings{li2017foveanet,
  title={Foveanet: Perspective-aware urban scene parsing},
  author={Li, Xin and Jie, Zequn and Wang, Wei and Liu, Changsong and Yang, Jimei and Shen, Xiaohui and Lin, Zhe and Chen, Qiang and Yan, Shuicheng and Feng, Jiashi},
  booktitle={ICCV},
  pages={784--792},
  year={2017}
}

@inproceedings{thavamani2021fovea,
  title={Fovea: Foveated image magnification for autonomous navigation},
  author={Thavamani, Chittesh and Li, Mengtian and Cebron, Nicolas and Ramanan, Deva},
  booktitle={ICCV},
  pages={15539--15548},
  year={2021}
}

@inproceedings{recasens2018learning,
  title={Learning to zoom: a saliency-based sampling layer for neural networks},
  author={Recasens, Adria and Kellnhofer, Petr and Stent, Simon and Matusik, Wojciech and Torralba, Antonio},
  booktitle={ECCV},
  pages={51--66},
  year={2018}
}

@article{achiam2023gpt,
  title={Gpt-4 technical report},
  author={Achiam, Josh and Adler, Steven and Agarwal, Sandhini and Ahmad, Lama and Akkaya, Ilge and Aleman, Florencia Leoni and Almeida, Diogo and Altenschmidt, Janko and Altman, Sam and Anadkat, Shyamal and others},
  journal={arXiv preprint arXiv:2303.08774},
  year={2023}
}

@article{gemini15,
  title={Gemini 1.5: Unlocking multimodal understanding across millions of tokens of context},
  author={Team, Gemini and Georgiev, Petko and Lei, Ving Ian and Burnell, Ryan and Bai, Libin and Gulati, Anmol and Tanzer, Garrett and Vincent, Damien and Pan, Zhufeng and Wang, Shibo and others},
  journal={arXiv preprint arXiv:2403.05530},
  year={2024}
}

@article{gemini,
  title={Gemini: a family of highly capable multimodal models},
  author={Team, Gemini and Anil, Rohan and Borgeaud, Sebastian and Alayrac, Jean-Baptiste and Yu, Jiahui and Soricut, Radu and Schalkwyk, Johan and Dai, Andrew M and Hauth, Anja and Millican, Katie and others},
  journal={arXiv preprint arXiv:2312.11805},
  year={2023}
}

@article{gpt3,
  title={GPT-3: Its nature, scope, limits, and consequences},
  author={Floridi, Luciano and Chiriatti, Massimo},
  journal={Minds and machines},
  year={2020},
  publisher={Springer}
}

@article{deepseek,
  title={Deepseek-v3 technical report},
  author={Liu, Aixin and Feng, Bei and Xue, Bing and Wang, Bingxuan and Wu, Bochao and Lu, Chengda and Zhao, Chenggang and Deng, Chengqi and Zhang, Chenyu and Ruan, Chong and others},
  journal={arXiv preprint arXiv:2412.19437},
  year={2024}
}

@article{llama,
  title={Llama: Open and efficient foundation language models},
  author={Touvron, Hugo and Lavril, Thibaut and Izacard, Gautier and Martinet, Xavier and Lachaux, Marie-Anne and Lacroix, Timoth{\'e}e and Rozi{\`e}re, Baptiste and Goyal, Naman and Hambro, Eric and Azhar, Faisal and others},
  journal={arXiv preprint arXiv:2302.13971},
  year={2023}
}

@article{llama2,
  title={Llama 2: Open foundation and fine-tuned chat models},
  author={Touvron, Hugo and Martin, Louis and Stone, Kevin and Albert, Peter and Almahairi, Amjad and Babaei, Yasmine and Bashlykov, Nikolay and Batra, Soumya and Bhargava, Prajjwal and Bhosale, Shruti and others},
  journal={arXiv preprint arXiv:2307.09288},
  year={2023}
}

@inproceedings{fan2024navigation,
  title={Navigation instruction generation with bev perception and large language models},
  author={Fan, Sheng and Liu, Rui and Wang, Wenguan and Yang, Yi},
  booktitle={ECCV},
  year={2024},
}

@inproceedings{wang2025g3d,
  title={g3d-lf: Generalizable 3d-language feature fields for embodied tasks},
  author={Wang, Zihan and Lee, Gim Hee},
  booktitle={CVPR},
  year={2025}
}

@article{hurst2024gpt,
  title={Gpt-4o system card},
  author={Hurst, Aaron and Lerer, Adam and Goucher, Adam P and Perelman, Adam and Ramesh, Aditya and Clark, Aidan and Ostrow, AJ and Welihinda, Akila and Hayes, Alan and Radford, Alec and others},
  journal={arXiv preprint arXiv:2410.21276},
  year={2024}
}
}

\iftoggle{arxiv}{%
    \clearpage
    \appendix

    \counterwithin{figure}{section}
    \counterwithin{table}{section}
    \counterwithin{equation}{section}
    \setcounter{table}{0}
    \setcounter{figure}{0}
    \setcounter{equation}{0}
    \section{Appendix}

    \let\origsection\section
    \let\origsubsection\subsection
    \let\origsubsubsection\subsubsection
    \let\section\subsection
    \let\subsection\subsubsection
  
    \begin{strip}
\centering
\resizebox{\linewidth}{!}{
\setlength{\tabcolsep}{6pt}
\begin{tabular}{lcccccccccc}
\toprule
\multirow{2}{*}{Benchmarks} 
& \multicolumn{4}{c}{Scan2Cap} 
& \multicolumn{5}{c}{ScanQA} 
& SQA3D \\

\cmidrule(lr){2-5} 
\cmidrule(lr){6-10} 
\cmidrule(l){11-11}

& CIDEr & BLEU-4 & METEOR & ROUGE 
& CIDEr & BLEU-4 & METEOR & ROUGE & EM 
& EM \\
\midrule
GPT-4o~\cite{hurst2024gpt} & 7.0 & 5.5 & 17.1 & 38.5 & 43.3 & 16.6 & 16.3 & 31.6 & 11.1 & 41.7 \\
\OURS{} & 90.5 & 41.7 & 28.9 & 62.2 & 107.1 & 17.8 & 20.8 & 50.9 & 31.2 & 62.3 \\
\bottomrule
\end{tabular}
}
\captionof{table}{Comparison between GPT-4o and \OURS{} across Scan2Cap, ScanQA, and SQA3D benchmarks.}
\label{tab:gpt}
\end{strip}

\section{Implementation Details}

\subsection{Evaluation with GPT Model}
To compare with strong closed-source systems, we evaluate GPT-4o~\cite{hurst2024gpt} on the benchmarks used in the main paper. Following the same strategy as \OURS{} and Video-3D-LLM~\cite{video3dllm}, we uniformly extract frames from the scene videos as input to generate visual tokens. Since GPT-4o produces only text output, it is largely unable to predict the 3D bounding-box coordinates required by the referring benchmarks (ScanRefer~\cite{scanrefer} and Multi3DRefer~\cite{multi3drefer}). Therefore, we report its performance only on Scan2Cap~\cite{scan2cap}, ScanQA~\cite{scanqa}, and SQA3D~\cite{sqa3d}. As shown in Table~\ref{tab:gpt}, without task-specific fine-tuning, even the powerful GPT-4o model falls short of the spatial understanding demonstrated by specialized models such as our proposed \OURS{}. This gap is particularly evident on the Scan2Cap~\cite{scan2cap} benchmark, where precise object localization based on coordinate inputs is essential.

\subsection{Task Relevant Objects}
We describe how target objects are identified and task-relevant object sets $\mathcal{E}_+$ are constructed from the five training datasets.

In ScanRefer~\cite{scanrefer}, each question is linked to one specific object from the given 3D scene. We directly use this object as the target object. 

For Multi3DRefer~\cite{multi3drefer}, each question may refer to zero, one, or multiple objects in the 3D scene. Consequently, the number of target objects for a sample from this dataset ranges from zero to several instances.

In Scan2Cap~\cite{scan2cap}, the model is tasked to give a precise description for a single referring object, with its 3D bounding box provided as input. Since the dataset provides the metadata of the object to be described for each data sample, we assign that object as the target.

ScanQA~\cite{scanqa} provides a large collection of QA pairs, each accompanied by an object ID list and a corresponding name list of all objects relevant to the pair. Since some samples include multiple objects from different categories and distant locations, which may impede the learning of our task-affinity token, we apply a filtering strategy. Specifically, we retain only cases where (1) the object list contains a single instance, or (2) all object names are identical and matches the answer, which commonly occurs in questions such as ``What is to the left of the chair?''.

SQA3D~\cite{sqa3d} benchmarks the situated Question Answering task in 3D Scenes. It does not provide corresponding objects to their QA pairs. Therefore, we do not compute the InfoNCE loss for the task-affinity token when training on its samples.

\subsection{\Grid{} Grid Prompt Template}
Given a set of 3D object bounding boxes, we retain their x- and y-axis coordinates and discretize them to place all objects into our \grid{} grid. The grid is first represented as a dictionary, where each key corresponds to a 2D grid cell coordinate $(x, y)$ and the value is the list of object names located in that cell. This dictionary is then used to populate the following text prompt:

\lstset{basicstyle=\small\ttfamily,breaklines=true,breakautoindent=false,breakindent=0pt}
\begin{lstlisting}[breaklines]
This is a top-down view of a scene divided into a {grid_H} by {grid_W} grid. Each cell
may contain multiple objects, and the objects are separated by commas. This is an
abstraction of the scene and might be incomplete.
At (row={x}, col={y}), there is: {obj_str},
At (row={x}, col={y}), there is: {obj_str},
...
\end{lstlisting}

}{}

\end{document}